%% file: paper.tex
\documentclass[]{fairmeta}

\usepackage{comment}
\usepackage{xspace}
\usepackage{marginnote}
\usepackage{multirow}
\usepackage{nicematrix}
\usepackage{wrapfig}
\usepackage{amsmath}
\usepackage{amssymb}
\usepackage{amsfonts}
\usepackage{algorithm}
\usepackage{algorithmicx}
\usepackage{algpseudocode}
\usepackage{float}
\usepackage{colortbl}
\usepackage{algorithm}
\usepackage{algpseudocode}

\usepackage{nicematrix}  
\usepackage{graphicx}
\usepackage{booktabs}
\usepackage{enumitem}

\definecolor{readableyellow}{RGB}{218, 165, 32}  

\makeatletter
\DeclareRobustCommand\onedot{\futurelet\@let@token\@onedot}
\def\@onedot{\ifx\@let@token.\else.\null\fi\xspace}

\definecolor{lightgray}{gray}{0.95}
\definecolor{color3}{gray}{0.95}
\definecolor{cvprblue}{rgb}{0.21,0.49,0.74}

\renewcommand{\paragraph}[1]{\vspace{.5em}\noindent\textbf{#1}}
\title{PhyGDPO: Physics-Aware Groupwise Direct Preference  Optimization for Physically Consistent Text-to-Video Generation}

\author[1,2,*]{Yuanhao Cai}
\author[1]{Kunpeng Li}
\author[3]{Menglin Jia}
\author[1]{Jialiang Wang}
\author[1]{Junzhe Sun}
\author[1]{Feng Liang}
\author[1]{Weifeng Chen}
\author[1]{Felix~Juefei-Xu}
\author[1]{Chu Wang}
\author[1]{Ali Thabet}
\author[1]{Xiaoliang Dai}
\author[4]{Xuan Ju}
\author[2,\dagger]{Alan Yuille}
\author[1,5,*\dagger]{Ji Hou}

\affiliation[1]{Meta Superintelligence Labs}
\affiliation[2]{Johns Hopkins University}
\affiliation[3]{Meta BizAI}
\affiliation[4]{CUHK}
\affiliation[5]{Tencent Hunyuan}

\contribution[*]{Work was done in Meta Superintelligence Labs}
\contribution[\dagger]{Corresponding authors}
\input{sec/00_abstract}

\usepackage{graphicx}
\usepackage{etoolbox}

\definecolor{resourcepink}{HTML}{D72A8A}

\renewcommand{\metadataformat}[2][]{%
    {\small\ifstrempty{#1}{}{{\sffamily\bfseries #1:}\ }#2}%
}

\newcommand{\resourceicon}[1]{%
    \raisebox{-0.15em}{%
        \includegraphics[height=0.95em]{fig/#1}%
    }%
}

\newcommand{\hficon}{%
    \raisebox{-0.25em}{%
        \includegraphics[height=1.35em]{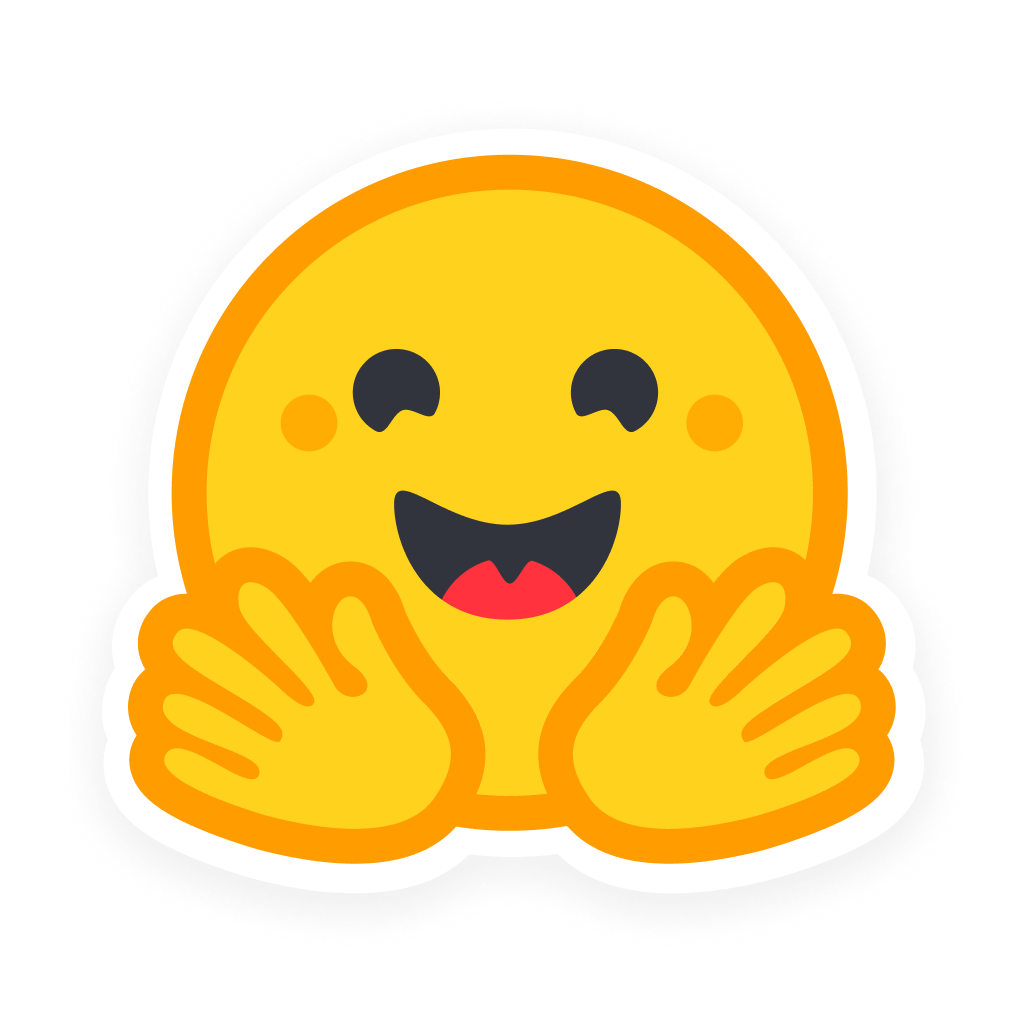}%
    }%
}

\DeclareRobustCommand{\phygdpolinks}{%
    {\fontsize{10}{12}\selectfont
    \hypersetup{urlcolor=resourcepink}%
    \mbox{%
        \href{https://caiyuanhao1998.github.io/project/PhyGDPO}
        {\resourceicon{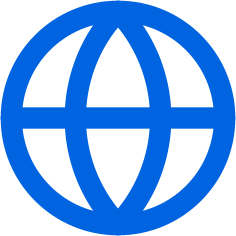}\,~Project Page}%
        ~~~~
        \href{https://github.com/caiyuanhao1998/Open-PhyGDPO}
        {\resourceicon{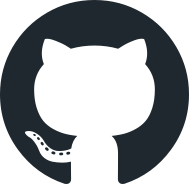}\,~Github Code}%
        ~~~~
        \href{https://huggingface.co/datasets/CaiYuanhao/PhyGDPO}
        {\hficon\,~Dataset}%
        ~~~~
        \href{https://huggingface.co/CaiYuanhao/PhyGDPO}
        {\hficon\,~Models}%
        ~~~~
        \href{https://huggingface.co/spaces/CaiYuanhao/PhyGDPO}
        {\hficon\,~Online Demo}%
    }}%
}

\metadata[]{\protect\raisebox{0pt}{\phygdpolinks}}

\begin{document}
\maketitle

\input{figures/fig_teaser}

\input{sec/01_introduction}
\input{sec/02_method}

\input{sec/04_experiments}
\input{sec/05_related_work}
\input{sec/06_conclusion}

\clearpage
\newpage
\bibliographystyle{assets/plainnat}
\bibliography{paper}

\input{sec/XX_supp}

\end{document}

%% file: sec/00_abstract.tex
\abstract{
Recent advances in text-to-video (T2V) generation have achieved good visual quality, yet synthesizing videos that faithfully follow physical laws remains an open challenge. Existing methods mainly based on graphics struggle to generalize beyond simple simulated environments. The scarcity of training data with rich physics interactions and phenomena is also a problem. In this paper, we first introduce a Physics-Augmented video data construction Pipeline, PhyAugPipe, that leverages a vision–language model (VLM) with chain-of-thought reasoning to collect a training dataset, PhyVidGen-135K. Then we formulate a principled Physics-aware Groupwise Direct Preference Optimization, PhyGDPO, framework that uses real-world video as winning case to guarantee correct physics learning and builds upon the groupwise Plackett–Luce probabilistic model to capture holistic preferences beyond pairwise comparisons. In PhyGDPO, we design a Physics-Guided Rewarding (PGR) scheme that leverages VLM-based physical rewards to direct the optimization to focus on challenging physics cases. In addition, we propose a LoRA-Switch Reference (LoRA-SR) scheme that avoids full-model duplication as reference for efficient DPO training. Experiments show that our method outperforms SOTA methods on PhyGenBench and VideoPhy2. The application of PhyGDPO on image-to-video generation also suggests the generalization ability of our algorithm.
}

%% file: figures/fig_teaser.tex
\begin{figure}[h!]
\centering
\vspace{3mm}
\includegraphics[width=\linewidth]{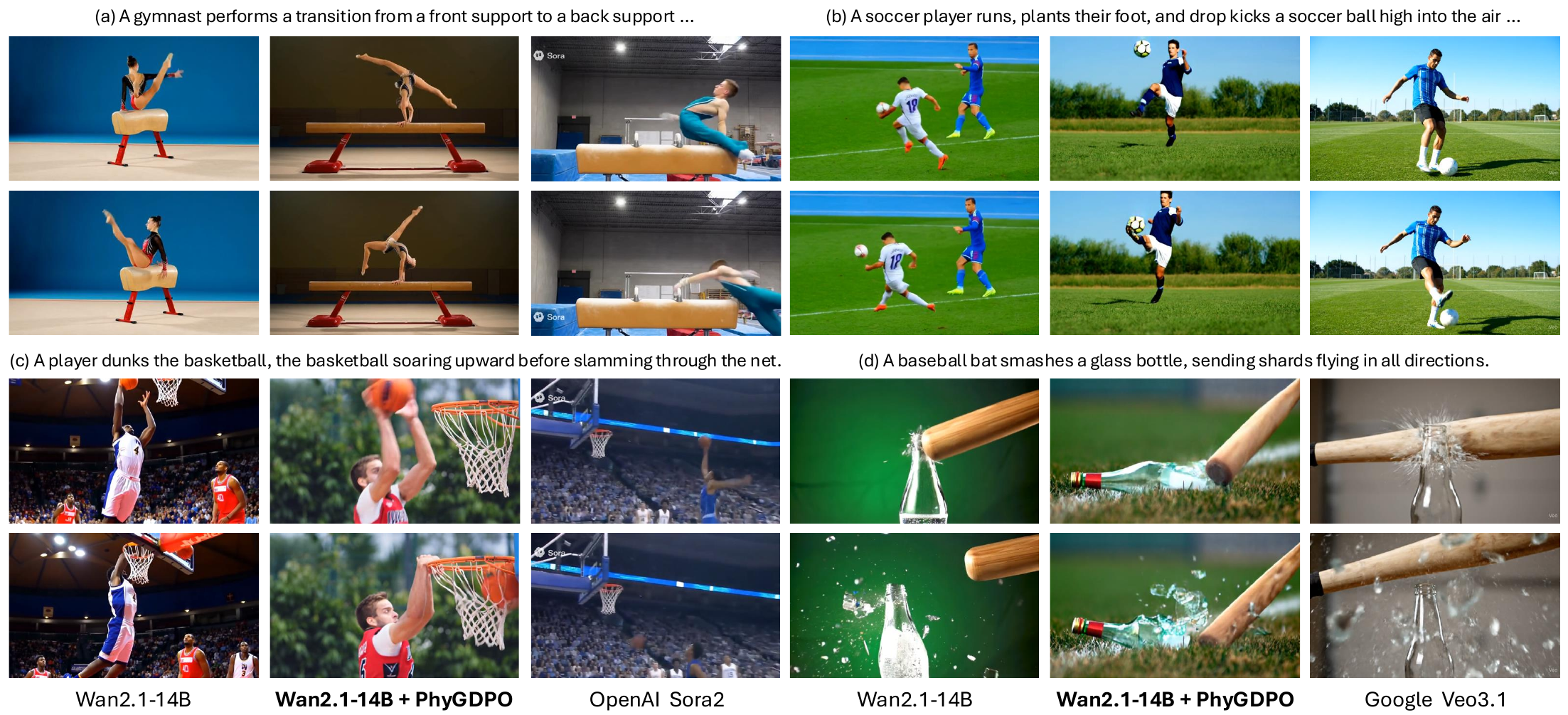}
\caption{
Text-to-video generation on four challenging action categories: (a) gymnastics, (b) soccer, (c) basketball, and (d) glass smashing. When using our post-training method, PhyGDPO, on Wan2.1-T2V-14B~\cite{wan}, the model yields more physically plausible results than OpenAI Sora2~\cite{sora} and Google Veo3.1~\cite{veo3} by generating well-structured human bodies and reasonable physical interactions such as foot kicking the soccer, basketball passing through the hoop, and glass shattering.
}
\label{fig:teaser}
\end{figure}

%% file: sec/01_introduction.tex
\vspace{-4mm}
\section{Introduction}
\label{section:intro}
\vspace{-2mm}

With the development of computing power and the increasing scale of training data, text-to-video (T2V) generation~\cite{wan,sora,videocrafter1} has witnessed significant progress. However, accurately and consistently modeling the physics in the generated video still remains challenging and less explored. Improving the physics reasoning capability of video generation models can make them closer to real-world simulators, which can benefit a wide range of applications such as video gaming~\cite{gamegen,GameNGen}, autonomous driving~\cite{drivedreamer,yang2024generalized,panacea}, robotics~\cite{cosmos,cosmos-transfer,chen2024igor}, film making~\cite{moviegen,cinemaster,moviebench}, \emph{etc}.

To improve the physics modeling, graphics-based methods~\cite{kang2024far} rely on simulation engines to specify physical parameters of simple scenarios such as perfectly elastic collisions and basic rigid body dynamics. Nonetheless, it is impractical to apply them in real-world scenes as the environments are too complex to parameterize. 

Another technical route~\cite{diffphy,phyt2v} is based on prompt extension with a large language model (LLM). These methods adopt an LLM to extend the prompts with explicit physics laws and phenomena, and then use the extended prompts as input to simulate the physics by iteratively generating the video or finetuning the model on a small subset. These methods simply follow the LLM-augmented prompts and lack the ability of thinking in physics. Instead, they use the LLM as their surrogate brain and outsource the reasoning process to it. Even so, the prompt following ability of current T2V models is still limited and current LLM’s physics reasoning ability is also weak and often erroneous, which may in turn mislead the T2V model when it follows such guidance to generate videos.


To learn the implicit physics, most current T2V foundation models are trained on massive collections of high-quality text–video data pairs. However, collecting and annotating such data is extremely costly and labor-intensive. Moreover, models obtained through such supervised fine-tuning (SFT) or training still exhibit limited physics reasoning ability. For instance, OpenAI Sora2~\cite{sora} and Google DeepMind Veo3.1~\cite{veo3} often fail in complex human motions or physical phenomena, as shown in Fig.~\ref{fig:teaser}, \ref{fig:comparison_1}, \ref{fig:comparison_2}, and \ref{fig:comparison_3}. The key reason is that there are no negative training data to provide contrastive signals discouraging physically inconsistent generations.

The emergence of direct preference optimization~\cite{dpo} (DPO) may provide a potential solution. However, directly using DPO may encounter three problems: (i) Lack of training data pairs that comprehensively capture physical activities, interactions, and phenomena. (ii) The correct physics preference optimization cannot be guaranteed because vanilla DPO uses generated video as winning case and the physical realism of the generated video is limited and some challenging cases remain difficult to generate with correct physics. The supervision is mainly based on the Bradley–Terry (BT) probabilistic model, which only compares a single pair of generated samples. Such single binary comparison struggles to capture inherently holistic global preference signals of physical plausibility.  
(iii) Vanilla DPO copies the full model as reference, which consumes substantial GPU memory and decreases  efficiency.

To address these issues, we firstly propose a data construction pipeline, PhyAugPipe, that exploits a vision-language model (VLM) to filter text-video data pairs capturing rich physical interactions and phenomena from a large T2V data pool by parsing the entities and reasoning their actions with our designed chain-of-thought (CoT)~\cite{cot} rule. We use PhyAugPipe to collect a training dataset, PhyVidGen-135K, containing 135K text-video pairs. Secondly, we formulate a novel Physics-aware Groupwise Direct Preference Optimization (PhyGDPO) framework for physically consistent T2V generation. PhyGDPO uses the real-world video that always follows physical laws as the winning case to guarantee correct physics learning. Different from vanilla pairwise DPO, PhyGDPO is based on the groupwise Plackett–Luce (PL) model, which captures the probability distribution over a group of candidate videos, enabling holistic preference adaptation beyond isolated pairwise comparisons. To further improve physics preference optimization, we propose a Physics-Guided Rewarding (PGR) scheme. PGR leverages a physics-aware VLM to guide data sampling and DPO training to focus on challenging actions and allows physics-violating samples to exert stronger influence. To improve DPO training efficiency and stability, we also propose a LoRA~\cite{hu2022lora}-Switch Reference (LoRA-SR) scheme that does not need to copy the full model as reference like vanilla DPO, which occupies redundant GPU memory. Instead, we freeze the base model as reference and attach LoRA with an environment manager to flexibly switch between reference and action modes. Benefit from our dataset and techniques, PhyGDPO boosts the physical plausibility of the base T2V model and yields better results than the closed-source models OpenAI Sora2~\cite{sora} and Google Veo3.1~\cite{veo3} on some challenging actions, as shown in Fig.~\ref{fig:teaser}, \ref{fig:comparison_1}, \ref{fig:comparison_2}, and \ref{fig:comparison_3}.

In a nutshell, our contributions can be summarized as follows:
\begin{itemize}
    \item We formulate a principled DPO framework, PhyGDPO, based on the groupwise Plackett-Luce probabilistic model to capture holistic physics advantage signal for physically consistent text-to-video generation.
    \vspace{1.5mm}
    \item We design PGR to guide data sampling and preference optimization to focus on challenging physics. We propose LoRA-SR scheme to reduce GPU memory occupancy for more efficient and stable DPO training.
    \vspace{1.5mm}
	\item We present PhyAugPipe to construct physics-rich text-video pairs. We use PhyAugPipe to collect a training dataset, PhyVidGen-135K, with over 135K data pairs for studying physically consistent T2V generation.
    \vspace{1.5mm}
    \item Experiments show that PhyGDPO surpasses SOTA methods on PhyGenBench and
VideoPhy2 while yielding higher human preference. The results on image-to-video generation also suggest the generality of our method.
\end{itemize}

%% file: sec/02_method.tex
\vspace{-10mm}
\section{Method}
\vspace{-1mm}
\label{section:model}

\begin{figure*}[t]
	\begin{center}
		\begin{tabular}[t]{c}  \hspace{-3.8mm}
\includegraphics[width=1.0\textwidth]{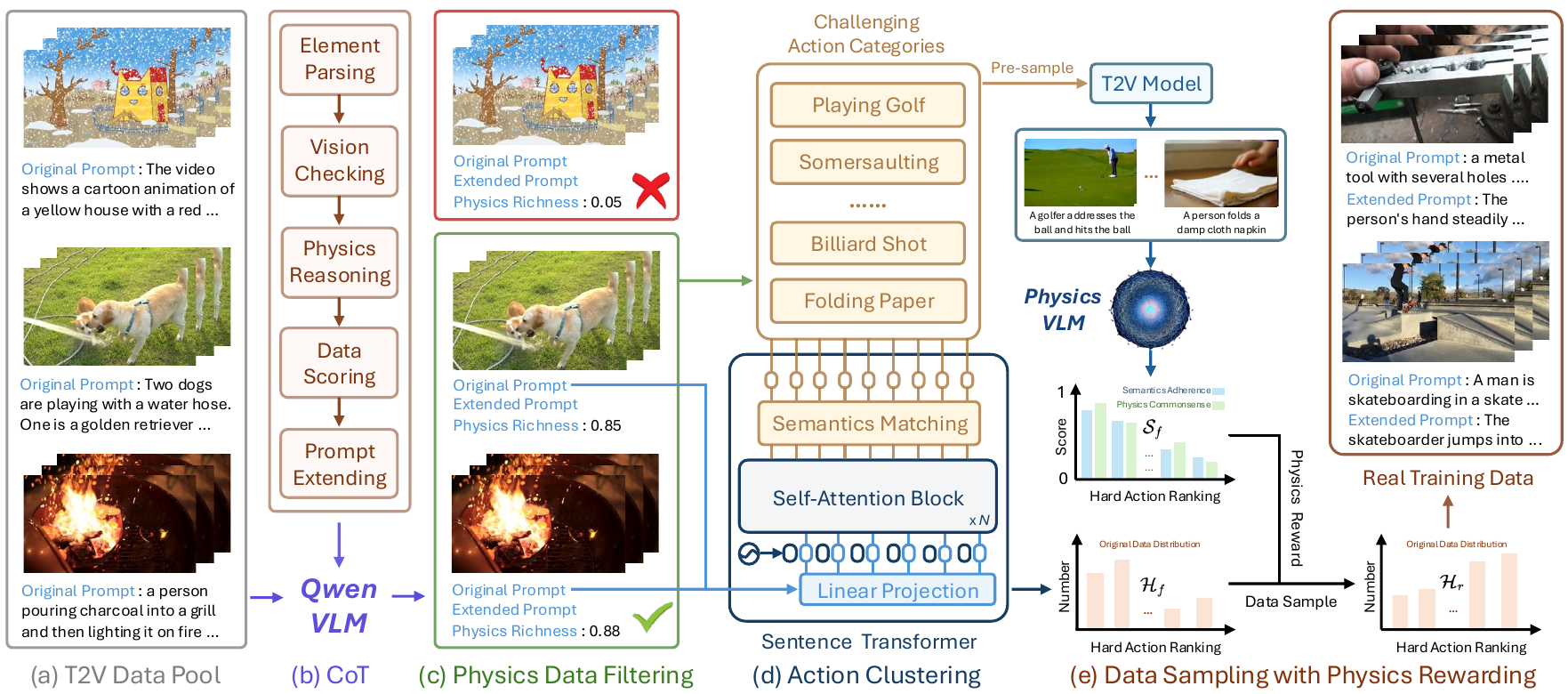}
		\end{tabular}
	\end{center}
	\vspace{0mm}
	\caption{\small Our physics-augmented video data construction pipeline (PhyAugPipe) first adopts a VLM, Qwen-2.5-72B-Instruct~\cite{qwen2}, following our designed chain-of-thought (CoT) rule in (b) to select text-video data pairs that contain rich physics interactions and phenomena from a large-scale high-quality text-video data pool in (a). Then in (d), we perform action clustering on the filtered data pairs from (c) through the semantics matching via a sentence Transformer~\cite{sentence_bert}. Subsequently, in (e), we adopt a physics-aware VLM, VideoCon-Physics~\cite{videophy}, to evaluate the difficulty of different action categories and then sample the text-video pairs accordingly as the wining cases of our training data for preference optimization.}
	\label{fig:data_pipeline}
	\vspace{0mm}
\end{figure*}

\subsection{Physics-Augmented Video Data Construction}
Due to the scarcity of text-video data pairs with rich physics interactions and phenomena, we design a Physics-Augmented video data construction Pipeline (PhyAugPipe). As shown in Fig.~\ref{fig:data_pipeline}, we first use a VLM, Qwen2.5-72B-Instruct~\cite{qwen2}, equipped with our chain-of-thought (CoT) reasoning in  (b) to derive physics-rich videos in (c) by filtering from a large T2V data pool in (a). Then in (d), we cluster the filtered data according to their action categories and sample the data with the rewards of a physics-aware VLM, VideoCon-Physics~\cite{videophy}, in (e).

\noindent\textbf{Data Filtering with CoT.} As shown in Fig.~\ref{fig:data_pipeline} (b),  Qwen2.5 with our CoT first parses the elements from the given prompts and video frames, including the physical objects with materials, actions and forces between them. Based on the parsed elements, Qwen2.5 reasons how the entities interact and what results, and rates the physics richness of the data with a score from 0 to 1. Eventually, Qwen2.5 extends the prompt with explicit physics reasoning. Although we explore the implicit physics reasoning ability of T2V models, but our dataset still include the extended prompts to support other research purposes, such as LLM-guided T2V. Please refer to Alg.~\ref{alg:cot_pipeline} for the details of our CoT rules. Then in Fig.~\ref{fig:data_pipeline} (c), we threshold the data according to the estimated physics richness.

\begin{algorithm}[t] 
\caption{Brief Pipeline of our Chain-of-Thoughts Prompts}
\label{alg:cot_pipeline}
\begin{algorithmic}[1]
\Require Original prompt $p$, video frames $V$
\Ensure JSON object containing: original, parse, reason, extended, physics\_richness, physics\_label

\Statex
\State \textbf{Step 1: Element Parsing}
\State Extract entities, actions, forces, and outcomes from $(p, V)$ using the VLM.
\State Ensure all agents are included and avoid unsupported or speculative items.
\State Save results into the dictionary \textit{parse}.

\Statex
\State \textbf{Step 2: Vision Checking}
\State Compare \textit{parse} with video frames and the original prompt.
\State Remove hallucinated elements and add missing visible entities or interactions.
\State Update \textit{parse} accordingly.

\Statex
\State \textbf{Step 3: Physics Reasoning}
\State Produce a concise explanation describing how the parsed entities interact through physical forces and lead to observed outcomes.
\State Store the explanation as \textit{reason}.

\Statex
\State \textbf{Step 4: Data Scoring}
\State Compute physics\_richness $\in [0,1]$ based on:
\Statex \quad -- number and interaction of entities;
\Statex \quad -- presence of explicit forces and outcomes;
\Statex \quad -- causal clarity in \textit{reason};
\Statex \quad -- penalties from camera motion, stylization, or static-aftermath filters.
\State Set physics\_label = 1 if physics\_richness $\ge 0.60$, otherwise $0$.

\Statex
\State \textbf{Step 5: Prompt Extending}
\State Extend the original prompt by inserting causal physical details based on \textit{reason}, without adding new entities, forces, or sensory descriptions.
\State Limit the extension to at most 100 words and store it as \textit{extended}.

\Statex
\State \Return JSON object containing all computed fields.

\end{algorithmic}
\end{algorithm}

\noindent\textbf{Action Clustering via Semantics Matching.} After data filtering, the retained samples may exhibit distributional imbalance on different actions. To check this issue, we categorize the filtered data into semantically coherent action clusters. Specifically, in Fig.~\ref{fig:data_pipeline} (d), we first compile a list containing $K_a$ challenging action categories. Then we feed the original text prompt of each filtered sample and the action list into a sentence Transformer~\cite{sentence_bert} to perform fuzzy semantics matching, thereby determining the action category to which each data sample belongs. We count the number of data samples in each action category to obtain the distribution histogram $\mathcal{H}_f$ for the filtered data.

\begin{figure*}[t]
	\begin{center}
		\begin{tabular}[t]{c}  \hspace{-6mm}
\includegraphics[width=1.03\textwidth]{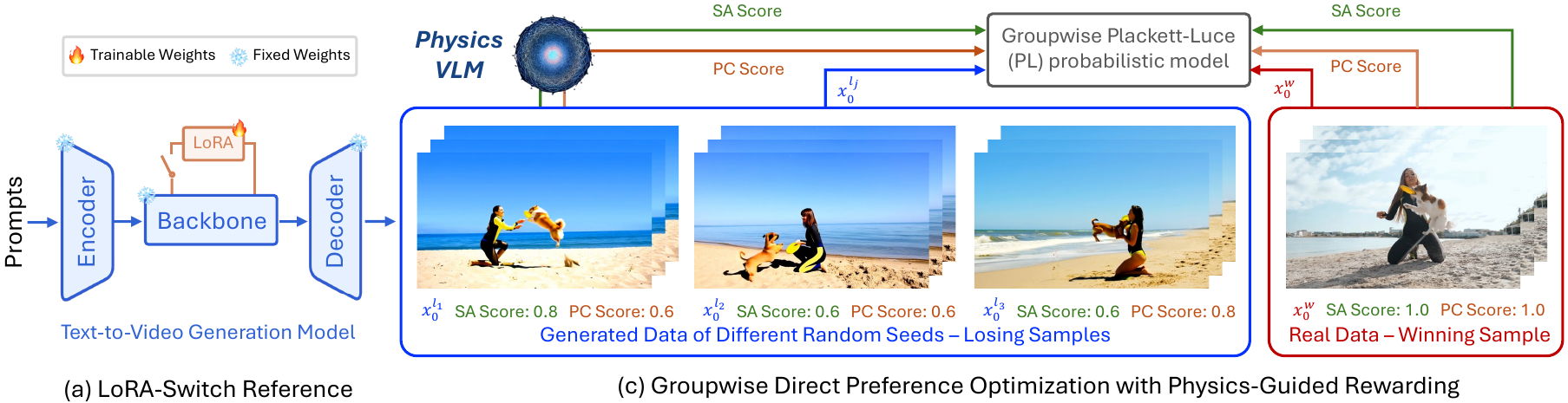}
		\end{tabular}
	\end{center}
	\vspace{1mm}
	\caption{\small Overview of PhyGDPO. (a) The LoRA-Switch Reference scheme can flexibly toggle between the reference and action modes for DPO advantage computation to save the GPU memory and increase the training stability. (b) our DPO framework is based on the groupwise Plackett-Luce (PL) probabilistic model and adopts a physics-aware VLM, VideoCon-Physics~\cite{videophy}, to reward the DPO training. The winning case is the real-world video because it always captures correct physics as a guarantee.}
	\label{fig:method}
	\vspace{1mm}
\end{figure*}

\noindent\textbf{Data Sampling with Physics Rewarding.} As different action categories vary significantly in physical complexity thus leading to different generation difficulties for the T2V model, we propose to balance the training data accordingly. In particular, we first evaluate how well the T2V model performs on each action category. Based on the performance distribution, we adjust the sampling ratio by allocating more samples to the categories where the model performs poorly. As shown in Fig.~\ref{fig:data_pipeline} (d), we first select the top-$n_c$ samples with the highest semantics matching scores within each category as its representatives. Then we employ a physics-aware VLM, VideoCon-Physics~\cite{videophy}, to evaluate each video representative. VideoCon-Physics outputs a semantics adherence score and a physics commonsense score $\in [0,1]$ to measure the physics plaucibility of the video. We average the two scores to obtain the overall score for each video. Then we compute the mean score of all representatives within each category to obtain the category score. In this way, we derive the score distribution histogram $\mathcal{S}_f$. Then we sample the data according to the difficulty of each action category. More specifically, we define the difficulty of the $k$-th action category as $d_k = 1 - \mathcal{S}_f(k)$ and assign a sampling weight following an exponential form as $r_k = \text{exp}(\tau d_k)$, where $\tau$ is a hyperparameter that controls how strongly the sampling favors difficult action categories. Given the total sampling budget $N$, the number of data pairs sampled from the $k$-th action category is $
    \mathcal{H}_r(k) = \min\bigg(\mathcal{H}_f(k),\, N \cdot \frac{r_k}{\sum_{j=1}^{K_a} r_j}\bigg),
$
where $\mathcal{H}_r$ denotes the distribution histogram after data sampling with physics rewarding. This sampling strategy allocates more data to challenging actions, encouraging the model to learn more complex physics during training.

Please note that our PhyAugPipe is a filtering-based data pipeline. The VLMs are used to select physics-rich scenes and hard actions from the real-world videos, which always follow the physical laws. Therefore, the VLM rewarding only impacts the physics richness rather than correctness. We use the original human-written prompts instead of the VLM rewritten ones to avoid the potential bias or errors of VLMs.

\subsection{Physics-Aware Groupwise Direct Preference Optimization}

Existing DPO algorithms are based on the Bradley–Terry (BT) probabilistic model, which compares pairwise data samples, showing limitations in capturing global and holistic preference signals and aligning with human feedback. Besides, vanilla DPO uses generated video as winning case but the physical realism of the generated video is limited and some challenging physics actions are difficult to generate. To address these issues, we formulate a Physics-Aware Groupwise Direct Preference Optimization (PhyGDPO), as illustrated in Fig.~\ref{fig:method}.

\noindent\textbf{Groupwise Probability.}
We denote the reward as $r(c,x_0)$ with the generation $x_0$ and condition $c$. Different from the normal DPO, we start from the groupwise Plackett-Luce (PL) probabilistic model, as shown in Fig.~\ref{fig:method} (b). We adopt the real video as the winning case $x_0^w$ because it always follows physical laws and a set of generated videos as the losing cases $\mathcal{G}^l(c)=\{x_0^{l_1},\ldots,x_0^{l_m}\}$. The preference probability of the PL model is formulated as
\begin{equation}
\small
p_{\mathrm{PL}}\!\big(x^{w}_0 \mid \mathcal{G}^l(c),c\big)
=\frac{\exp\!\big(r(c,x^{w}_0)\big)}
{\exp\!\big(r(c,x^{w}_0)\big)+\sum_{j=1}^{m}\exp\!\big(r(c,x^{l_j}_0)\big)}.
\end{equation}
$r(c,x_0)$ can be parameterized by a neural network $\phi$~\cite{dpo} and estimated via maximum likelihood training as
\begin{equation}
\normalsize
\mathcal{L}_{\mathrm{PL}}(\phi)
= -\,\mathbb{E}_{c,\,\mathcal{G}^l(c)}\!
\Bigg[
\log
\frac{\exp\!\big(r_\phi(c,x^{w}_0)\big)}
{\exp\!\big(r_\phi(c,x^{w}_0)\big)+\sum_{j=1}^{m}\exp\!\big(r_\phi(c,x^{l_j}_0)\big)}
\Bigg].
\label{eq:l_pl}
\end{equation}
Vanilla DPO~\cite{videodpo,diffusiondpo} represents $r(c,x_0)$ by the T2V model itself as
\begin{equation}
\normalsize
r(c,x_0)
= \beta \log\frac{p_\theta^*(x_0|c)}{p_{\psi}(x_0|c)}
+ \beta \log Z(c),
\label{eq:reward}
\end{equation}
where $p_\theta$ and $p_{\psi}$ denote the conditional probabilities of trained model $\theta$ and reference model $\psi$. $p_\theta^*$ and $Z(c)$ are the unique global optimal solution and the partition function. 
We plug Eq.~\eqref{eq:reward} into Eq.~\eqref{eq:l_pl} and drop the group-constant term $\beta\log Z(c)$ as it does not affect the optimization direction to obtain the groupwise DPO (GDPO) loss as
\begin{equation}
\normalsize
\begin{aligned}
\mathcal{L}_{\text{GDPO}}(\theta)
&= -\,\mathbb{E}_{c,\,\mathcal{G}^l(c)}
\Big[
\log
\frac{\exp\!\big(\beta f_\theta(x^{w}_0,c)\big)}
{\exp\!\big(\beta f_\theta(x^{w}_0,c)\big)+\sum_{j=1}^{m}\exp\!\big(\beta f_\theta(x^{l_j}_0,c)\big)}
\Big] \\[4pt]
&= \mathbb{E}_{c,\,\mathcal{G}^l(c)}
\Big[
\log
\frac{\exp\!\big(\beta f_\theta(x^{w}_0,c)\big)+\sum_{j=1}^{m}\exp\!\big(\beta f_\theta(x^{l_j}_0,c)\big)}
{\exp\!\big(\beta f_\theta(x^{w}_0,c)\big)}
\Big] \\[4pt]   
&= \mathbb{E}_{c,\,\mathcal{G}^l(c)}
\Big[
\log\!\Big(1+\sum_{j=1}^{m}\exp\!\big(\beta(f_\theta(x^{l_j}_0,c)-f_\theta(x^{w}_0,c))\big)\Big)
\Big].
\end{aligned}
\label{eq:L_gdpo}
\end{equation}

\begin{figure*}[t]
	\begin{center}
		\begin{tabular}[t]{c}  \hspace{-4mm}
\includegraphics[width=1\textwidth]{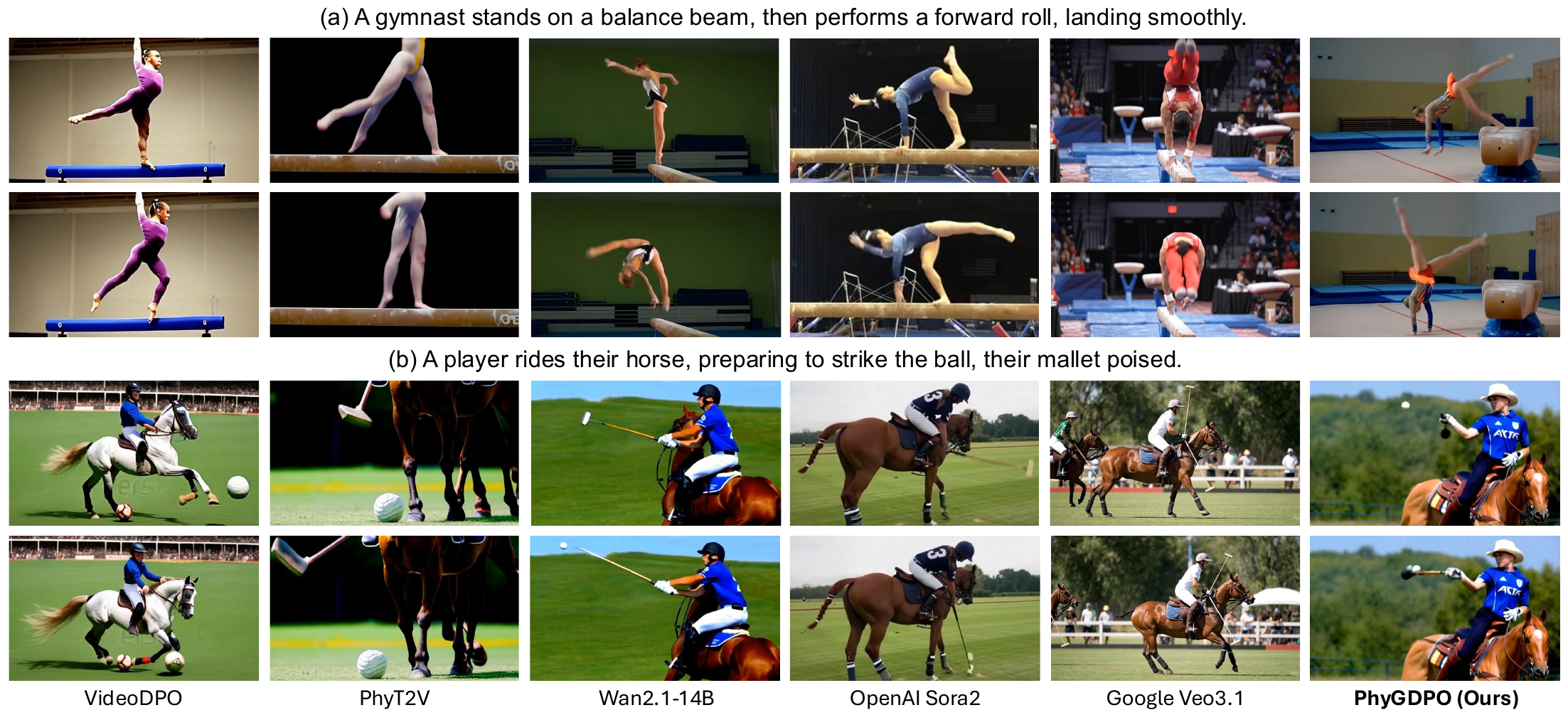}
		\end{tabular}
	\end{center}
	\vspace{0mm}
	\caption{\small Results of two challenging actions (gymnastics and polo) on VideoPhy2. Our method generates more physically consistent videos, showing coherent, deformation-free gymnastic movements and realistic ball–mallet striking interactions.}
	\label{fig:comparison_1}
\end{figure*}

where we define the function $f_\theta(x_0,c)$ as the  log-likelihood ratio as
\begin{equation}
\normalsize
    f_\theta(x_0,c) = \log\frac{p_\theta(x_0|c)}{p_{\psi}(x_0|c)} = \sum_{k=1}^T \log \frac{p_\theta(x_{t_{k-1}} \mid x_{t_k}, c)}{p_{\psi}(x_{t_{k-1}} \mid x_{t_k}, c)}
\;\triangleq\; \sum_{k=1}^T \Delta_k.
\end{equation}
Here the terminal distribution is inherently represented as the joint transition process over multiple timesteps $t_k$ by the definition of diffusion or flow matching models.
Subsequently, we reformulate Eq.~\eqref{eq:L_gdpo} and estimate its upper bound using the Jensen’s inequality into a single timestep for more efficient training as
\begin{equation}
\normalsize
\begin{aligned}
    \mathcal{L}_{\text{GDPO}}(\theta) 
& = \mathbb{E}_{c,\,\mathcal{G}^l(c)}
\Big[
\log\!\Big(1+\sum_{j=1}^{m}\exp\!\big(\beta T\mathbb{E}_k[\Delta_k^{l_j} - \Delta_k^{w}]\big)\Big)
\Big] \\
& \leq \mathbb{E}_{c,\mathcal{G}^l(c),k}
\Big[
\log\!\Big(1+\sum_{j=1}^{m}\exp\!\big(\beta T[\Delta_k^{l_j} - \Delta_k^{w}]\big)\Big)
\Big].
\end{aligned}
\label{eq:loss_single_step}
\end{equation}
Although the upper bound in Eq.~\eqref{eq:loss_single_step} allows single-timestep training, it needs to infer the  model $2m+2$ times for each group, which requires a long time and large GPU memory. To handle this issue, we exploit an inequality as
\begin{equation}
\normalsize
1 + \sum_{j=1}^m e^{x_j} \leq
\prod_{j=1}^m \big(1+e^{\alpha_j x_j}\big)^{\gamma_j}, ~
 ~~
0 < \alpha_j \leq 1,\ \gamma_j \geq 1/\alpha_j.
\label{eq:prod-ineq}
\end{equation}
Please refer to the supplementary for the detailed proof process of this inequality. Subsequently, we can further formulate the upper bound of $\mathcal{L}_{\text{GDPO}}(\theta)$ via Ineq.~\eqref{eq:prod-ineq} into a single sample within each group for efficient training as
\begin{equation}
\normalsize
\begin{aligned}
    \mathcal{L}_{\text{GDPO}}(\theta) &\leq \mathbb{E}_{c,\mathcal{G}^l(c),k}
\Big[
\sum_{j=1}^{m}\gamma_j\log\Big(1+\exp\big({-\alpha_j\beta T[\Delta_k^{w} - \Delta_k^{l_j}]})\big)
\Big] \\
&= \mathbb{E}_{c,\mathcal{G}^l(c),k,j}
\Big[- m \gamma_j \log \sigma\big(\alpha_j \beta T (\Delta_k^{w} - \Delta_k^{l_j})\big)\Big]. \\
\end{aligned}
\label{eq:loss_single_step_2}
\end{equation}
This upper bound allows efficient training with single data pair sampling in a single timestep for each iteration. 

\begin{figure*}[t]
	\begin{center}
		\begin{tabular}[t]{c}  \hspace{-3.5mm}
\includegraphics[width=1\textwidth]{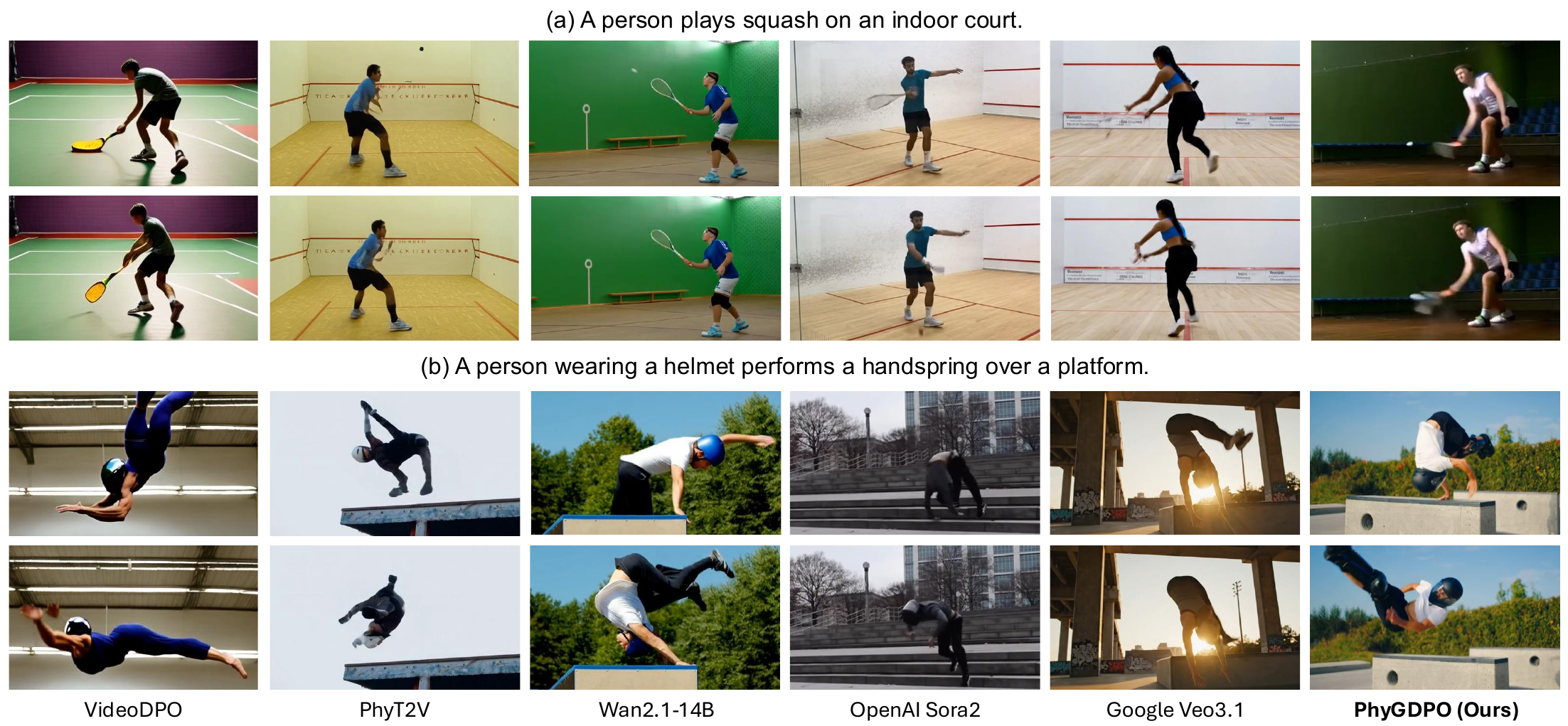}
		\end{tabular}
	\end{center}
	\vspace{0mm}
	\caption{\small Comparison on two challenging random user-input actions (squash and handspring). Our method generates more physically plausible videos, capturin racket–ball interactions in squash and well-coordinated body motion in handspring.}
	\label{fig:comparison_2}
\end{figure*}

\noindent\textbf{Physics-Guided Rewarding.}  To further improve physics preference optimization, we design a Physics-Guided Rewarding (PGR) to direct DPO training to focus on challenging physics cases. More specifically, we parameterize $\gamma_j$ and $\alpha_j$ in Eq.~\eqref{eq:loss_single_step_2} by the normalized semantics adherence score $s^{sa}_j \in [0, 1]$ and physics commonsense score $s^{pc}_j \in [0, 1]$ predicted by a physics-aware VLM, VideoCon-Physics, as
\vspace{1mm}
\begin{equation}
\normalsize
\begin{aligned}
v_j &= 1-\frac{s^{sa}_j + s^{pc}_j}{2}, 
~~\gamma_j = \frac{1 + \lambda \cdot \sigma\!\big(\kappa_\gamma (v_j - b_\gamma)\big)}{\alpha_{\text{min}}}, \\
\alpha_j &= \alpha_{\text{min}} + (1 - \alpha_{\text{min}}) \cdot \sigma\big(\kappa_\alpha (v_j - b_\alpha)\big),
\label{eq:alpha_gamma}
\end{aligned}
\vspace{1mm}
\end{equation}
where $v_j$ measures the physics difficulty and mainly modulates  $\gamma_j$ and $\alpha_j$, $\alpha_{\text{min}}$ sets a lower bound to avoid vanishing gradients, and the sigmoid function $\sigma(\cdot)$ ensures a bounded and smooth adjustment of $\gamma_j$, stabilizing the optimization. $\alpha_j$ adaptively adjusts the sharpness of the preference comparison, while $\gamma_j$ amplifies the physics-guided reward for samples with lower physical plausibility. They are controlled by the hyperparameters $\lambda$, $\kappa_{\alpha,\gamma}$,  $b_{\alpha,\gamma}$, and $\alpha_{\text{min}}$. Our designed physics rewards $(\alpha_j, \gamma_j)$ balances the training stability with adaptivity, allowing physics-violating samples to exert stronger influence during optimization. Please note that the VLM rewarding only affects the importance of samples while the correctness of physical supervision is guaranteed by the core objective, \emph{i.e.}, the DPO advantage of winning case (real-world video capturing correct physics) over losing case.

\noindent\textbf{Flow Matching Probabilistic Modeling.} Our method is based on the standard rectified flow matching as
\begin{equation}
\normalsize
x_t = (1-t)\,x_0+t\,x_1, ~~ x_1\sim\mathcal{N}(0,I),
\end{equation}
which induces the oracle velocity $
u^*(x_t,t\mid c)=x_1-x_0.
$
Discretizing time with step size $h$ (so $t\mapsto t-h$), the backward update is $x_{t-h}=x_t-h\,u^*(x_t,t\mid c)$. Since the rectified flow one-step reverse transition has no intermediate noise, we follow a vanishing-noise regularization and approximate $p_\theta(x_{t-h}\mid x_t,c)$ and $p_{\psi}(x_{t-h}\mid x_t,c)$ as
\vspace{1mm}
\begin{equation}
\normalsize
\begin{aligned}
p_\theta(x_{t-h}\mid x_t,c)&\approx \mathcal{N}\!\big(x_t-h\,v_\theta(x_t,t,c),\,\varepsilon I\big),\\
p_{\psi}(x_{t-h}\mid x_t,c)&\approx \mathcal{N}\!\big(x_t-h\,v_{\psi}(x_t,t,c),\,\varepsilon I\big),
\vspace{1mm}
\end{aligned}
\end{equation}
then let $\varepsilon\to0$ and use the log-ratio between Gaussians with the identity $(x_{t-h}-x_t)/h=-u^*(x_t,t\mid c)$, we obtain
\vspace{1mm}
\begin{equation}
\label{eq:rf-logratio}
\begin{aligned}
\log \frac{p_\theta(x_{t-h}\mid x_t,c)}{p_{\psi}(x_{t-h}\mid x_t,c)}
&\approx
-\frac{h^2}{2\varepsilon}\Big(\ell_\theta(x_t,t) - \ell_{\psi}(x_t,t)\Big), \\
\ell_\theta(x_t,t) &= \|v_\theta(x_t,t,c)-u^*(x_t,t\mid c)\|_2^2, \\
\ell_{\psi}(x_t,t) &= \|v_{\psi}(x_t,t,c)-u^*(x_t,t\mid c)\|_2^2,
\end{aligned}
\vspace{1mm}
\end{equation}
and reformulate the upper bound of Eq.~\eqref{eq:loss_single_step_2} into the final overall training objective as
\vspace{1mm}
\begin{equation}
\label{eq:rf-fmdpo}
\normalsize
\mathcal{L} =
\mathbb{E}_{c,\mathcal{G}^l(c),k,j} \Big[-m \gamma_j
\log \sigma\!\big(
-\alpha_j\beta T
[
(\ell_\theta^w-\ell_{\psi}^w)
-
(\ell_\theta^{l_j}-\ell_{\psi}^{l_j})
]
\big)\Big],
\vspace{1mm}
\end{equation}
where any global constant scale factor such as $h^2 / 2\varepsilon$ in Eq.~\eqref{eq:rf-logratio} can be absorbed into the hyperparameter $\beta$. For simplicity, we denote $\ell_\theta^w = \ell_\theta(x_{t_k}^w \mid t_k)$,
$\ell_{\psi}^w = \ell_{\psi}(x_{t_k}^w \mid t_k)$,
$\ell_\theta^{l_j} = \ell_\theta(x_{t_k}^{l_j} \mid t_k)$,
and $\ell_{\psi}^{l_j} = \ell_{\psi}(x_{t_k}^{l_j} \mid t_k)$ in Eq.~\eqref{eq:rf-fmdpo}.

\noindent\textbf{LoRA-Switch Reference.} 
Previous DPO methods usually copy the full model and fix it as the reference, which occupies redundant GPU memory, degrades computational efficiency, and hinders the scalability of model size. Plus, this full-copy strategy often leads to unstable and less effective DPO training because it updates the entire set of the weights and may cause the action model to quickly and dramatically deviate from the reference model. 

\begin{table*}[t]\hspace{-2.5mm}
    \subfloat[\small Quantitative results on VideoPhy2 \label{tab:videophy2}]{
        \renewcommand{\arraystretch}{1.3}
		\scalebox{0.57}{\begin{tabular}{lcccc}
			\toprule[0.2em]
			\rowcolor{lightgray}
			Methods~~
			& Hard
			& Activity
			& Interaction
			& Overall
			\\
			\midrule
			Vcrafter2 \cite{videocrafter2}
            & 0.0222 &0.1071 &0.0943 & 0.1034
            \\
            Wan2.1-14B \cite{wan}
            &0.0111 &0.1190 & 0.1572 &0.1288
            			\\
            Hunyuan \cite{hunyuan}
            & 0.0222 & 0.1286 & 0.1572 & 0.1356
            \\
            VideoDPO \cite{videodpo}
            &0.0167 &0.1310 &0.1572 & 0.1373
            \\
            PhyT2V \cite{phyt2v} & 0.0389 & 0.1405 & 0.1698 & 0.1492
            \\
            Sora2 \cite{sora}
            & 0.0389 & 0.1429 & 0.1698 & 0.1508
            \\
            Veo3.1 \cite{veo3}
            & 0.0444 & 0.1405 & 0.1887 & 0.1525
            \\
			PhyGDPO (Ours)
			& 0.0500  & 0.1571  &0.1761  & 0.1627
			\\
			\bottomrule[0.2em]
		\end{tabular}}}\hspace{0mm}
    \subfloat[\small Quantitative results on PhyGenBench\label{tab:phygenbench}]{
        \renewcommand{\arraystretch}{1.3}
		\scalebox{0.57}{\begin{tabular}{lcccccc}
			\toprule[0.2em]
			\rowcolor{lightgray}
			Methods
			& Mechanics
			& Optics
			& Thermal
			& Material
			& Avg
			\\
            \midrule
			Lavie \cite{lavie}
			& 0.40 & 0.44 & 0.38 & 0.32 & 0.36
			\\
			Wan2.1-14B \cite{wan}
			& 0.36 & 0.53 & 0.36 & 0.33 & 0.40
			\\
			Open-Sora \cite{opensora}
			& 0.43 & 0.50 & 0.44 & 0.37 & 0.44
			\\
			Pika \cite{pika}
			&0.35 &0.56 &0.43 &0.39 &0.44
            \\
			Vchitect-2.0 \cite{vchitect}
			& 0.41 & 0.56 & 0.44 & 0.37 & 0.45
			\\
            PhyT2V \cite{phyt2v}
            & 0.45 & 0.55 & 0.43 & 0.53 & 0.50
            \\
            VideoDPO \cite{videodpo}
            &0.48 & 0.60 &0.47 & 0.58 &0.54
			\\
			PhyGDPO (Ours)
			& 0.55  & 0.60  & 0.58  &0.47  & 0.55
			\\
			\bottomrule[0.2em]
		\end{tabular}}}\hspace{0mm}
    \subfloat[\small User preference of our method \label{tab:user_study}]{
        \renewcommand{\arraystretch}{1.3}
		\scalebox{0.57}{\begin{tabular}{lc}
			\toprule[0.2em]
			\rowcolor{lightgray}
			Compared Methods~~
            & ~~~Preference of Ours~~~
			\\
			\midrule
			Vcrafter2 \cite{videocrafter2}
            & 94.2 \%
            \\
            VideoDPO \cite{videodpo}
            &89.4 \%
            \\
            PhyT2V \cite{phyt2v}
            & 88.5 \%
            \\
            Wan2.1-14B \cite{wan} & 86.5 \%
            \\
            Hunyuan \cite{hunyuan}
            & 82.7 \%
            \\
            Sora2 \cite{sora}
            & 67.3 \%
            \\
            Veo3.1 \cite{veo3}
            & 64.4 \%
            \\
			PhyGDPO (Ours) & --
			\\
			\bottomrule[0.2em]
		\end{tabular}}} \vspace{-1mm}
	\caption{\small Quantitative comparison and user preference (\%) of our method on VideoPhy2~\cite{videophy2} and PhyGenBench~\cite{phygenbench} datasets.}
    \vspace{0mm}
	\label{tab:main_physics}
\end{table*}

\begin{figure*}[t]
	\begin{center}
		\begin{tabular}[t]{c}  \hspace{-4mm}
\includegraphics[width=1.01\textwidth]{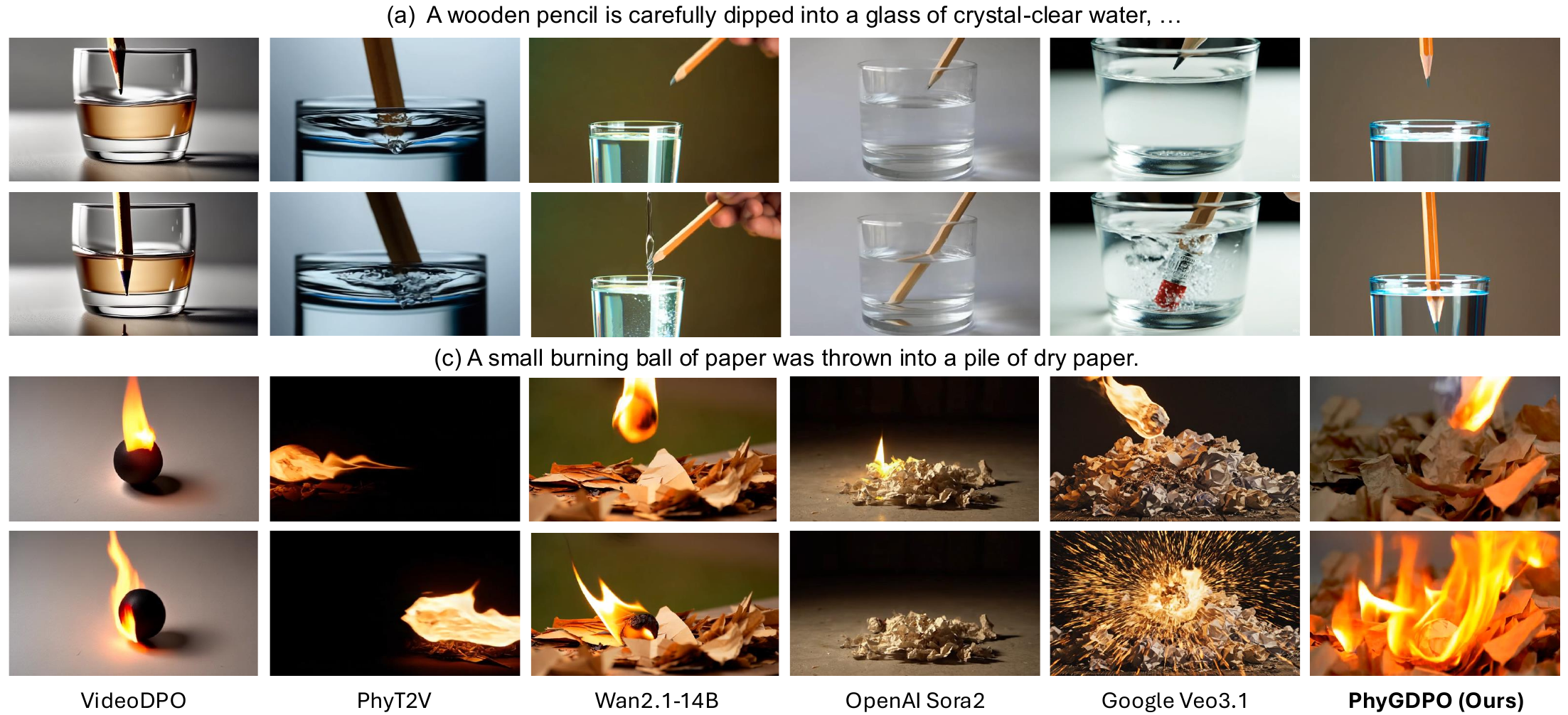}
		\end{tabular}
	\end{center}
	\vspace{1mm}
	\caption{\small Results of two challenging physics phenomena (water refraction and paper combustion) on PhyGenBench~\cite{phygenbench}. Our method can model the magnifying convex-lens effect and the realistic light refraction caused by the curved water surface, as well as the physically correct ignition and flame propagation on paper, demonstrating stronger physics reasoning ability.}
	\label{fig:comparison_3}
	\vspace{1mm}
\end{figure*}

To address this issue, we design a LoRA-Switch Reference (LoRA-SR) scheme. As shown in Fig.~\ref{fig:method} (b), we freeze the backbone as the reference model $\psi$ and attach trainable LoRA modules to $\psi$ as the action model $\theta$. Specifically, we attach LoRA to the query, key, value, and output linear projection layers of each self-attention in the Transformer backbone of the T2V model and implement an environment manager as the switch to flexibly toggle between the reference and action modes. Given the input tokens of the linear projection layer where the LoRA is attached to as $\mathbf{X} \in \mathbb{R}^{B\times L\times C_{in}}$, the output tokens $\mathbf{Y} \in \mathbb{R}^{B\times L\times C_{in}}$ of our LoRA-SR scheme can be formulated as
\begin{equation}
\small
    \mathbf{Y} = \mathbf{X} (\mathbf{W} + \mathbf{1}_{action} \cdot \Delta\mathbf{W})^{\top} = \mathbf{X} (\mathbf{W} + \mathbf{1}_{action} \cdot \frac{\alpha}{r} \mathbf{B} \mathbf{A})^{\top},
\end{equation}
where $\mathbf{W} \in \mathbb{R}^{C_{out} \times C_{in}}$ is the frozen weight of the linear layer,
$\mathbf{1}_{action} \in \{0,1\}$ is a binary indicator governed by the environment manager to switch between the reference ($\mathbf{1}_{action}=0$) and action ($\mathbf{1}_{action}=1$) modes,
$\alpha$ is the scaling factor,
$r$ is the rank,
and $\mathbf{B} \in \mathbb{R}^{C_{out} \times r}$ and 
$\mathbf{A} \in \mathbb{R}^{r \times C_{in}}$ are the learnable LoRA parameters. Our LoRA-SR enables the reference and trainable models to share the same heavy backbone while flexibly switching the lightweight LoRA parameters for the action and reference models. Thus, we can compute the DPO advantage without storing or loading an extra full model. This significantly increases computational efficiency, improves the scalability of model size, and prevents the model from drifting too far from reference, thus stabilizing the training process.

%% file: sec/04_experiments.tex
\section{Experiment}
\label{sec:experiment}

\noindent\textbf{Dataset.} We first use PhyAugPipe to process one million high-quality text-video pairs. After physics data filtering in Fig.~\ref{fig:data_pipeline} (c), we obtain 135K data pairs with sufficient physics interaction and phenomena. Then we perform action clustering and data resampling with physics rewarding to derive 17K data pairs from the 135K filtered data samples. Subsequently, we use the pre-trained T2V model Wan2.1-14B to generate videos for the 17K text prompts with different random seeds and use VideoCon-Physics~\cite{videophy} to score them. We adopt the short unextended prompts from the two datasets, VideoPhy2~\cite{videophy2} and PhyGenBench~\cite{phygenbench}, for evaluation. 

\noindent\textbf{Evaluation Metrics.} VideoPhy2 uses another VLM - VideoPhy2-AutoRater~\cite{videophy2}, which is different from VideoCon-Physics~\cite{videophy}, to evaluate the semantics adherence and physics commonsense of the videos with scores ranging from 1 to 5, and then compute the ratio with both scores $\geq$ 4. PhyGenBench evaluation is based on GPT-4o~\cite{gpt4o}, CLIP~\cite{clip}, InternVideo2~\cite{internvideo2}, and VideoLLaVA~\cite{videollava}. It firstly detects key physical phenomena required by the prompt and then assesses their order and naturalness, covering 27 physical laws in 4 basic physics domains.

\begin{table*}[t]
	\subfloat[\small Break-down ablation of PhyAugPipe on Wan2.1-14B\label{tab:phyaugpipe}]{\hspace{-3mm}
            \renewcommand{\arraystretch}{1.13}
		\scalebox{0.64}{\begin{tabular}{l c c c c}
				\toprule[0.15em]
				\rowcolor{color3}Method &~~~Hard~~~ &~~~~Activity~~~~ &~~~Interaction~~~  &~~Overall~~ \\
				\midrule[0.1em]
                    Baseline & 0.0333 &0.1405 &0.1635 &0.1475 \\
				+ Chain-of-Thought &0.0389 &0.1452 &0.1698 &0.1525 \\
				+ Action Clustering &0.0500 &0.1524 &0.1698 &0.1575\\
                + Physics Rewarding & 0.0500  & 0.1571  &0.1761  & 0.1627\\
				\bottomrule[0.15em]
	\end{tabular}}} \vspace{-1.3mm}\hspace{-1mm}
	\subfloat[\small Break-down ablation study of PhyGDPO on Wan2.1-14B  \label{tab:phygdpo}]{\hspace{-0.5mm}
            \renewcommand{\arraystretch}{1.13}
		\scalebox{0.64}{\begin{tabular}{l c c c c}
				\toprule[0.15em]
				\rowcolor{color3}Method &~~~Hard~~~ &~~~Activity~~~ &~~~Interaction~~~  &~~Overall~~ \\
				\midrule[0.1em]
                Baseline &0.0111 &0.1190 & 0.1572 &0.1288 \\
                + LoRA-SR &0.0278 &0.1381 &0.1635 &0.1458 \\
                    + Groupwise Model&0.0389 &0.1500 &0.1698 &0.1559 \\
				+ Physics Rewarding & 0.0500  & 0.1571  &0.1761  & 0.1627 \\
				\bottomrule[0.15em]
	\end{tabular}}} \vspace{-1.3mm}
	\subfloat[\small Comparison with SOTA DPO methods on Wan2.1-1.3B \label{tab:dpo_compare}]{\hspace{-3mm}
            \renewcommand{\arraystretch}{1.13}
		\scalebox{0.645}{\begin{tabular}{l c c c c c}
				\toprule[0.15em]
				\rowcolor{color3} Method &~~~~~Hard~~~~~ &~~~~~Activity~~~~~  &~~~~Interaction~~~~ &~~Overall~~ \\
				\midrule[0.1em]
                Baseline   &0.0118 &0.1136 &0.1447 &0.1232 \\
                Flow-DPO~\cite{flowdpo}   &0.0296 &0.1247
 &0.1342 &0.1283 \\
                   VideoDPO~\cite{videodpo}   &0.0278 &0.1262
 &0.1509 &0.1305 \\
				  PhyGDPO (Ours)&0.0444 &0.1357 &0.1635 &0.1407 \\
				\bottomrule[0.15em]
	\end{tabular}}}\hspace{0.5mm}\vspace{0.5mm}
	\subfloat[\small Ablation of LoRA-SR mechanism on Wan2.1-1.3B\label{tab:lora_sr}]{
            \renewcommand{\arraystretch}{1.13}
		\scalebox{0.645}{\begin{tabular}{l c c c c}
				\toprule[0.15em]
				\rowcolor{color3}  Method & ~GPU Memory~ &Storage Space & ~Hard Score~ & Overall Score\\
				\midrule[0.1em]
                Baseline   &- &- &0.0118 &0.1232 \\
                LoRA-SFT   &24.7GB &84MB &0.0167 &0.1283 \\
                    w/o LoRA-SR &48.7GB &5.3GB &0.0389 &0.1373 \\
				  with LoRA-SR &25.3GB &84MB &0.0444 &0.1407 \\
				\bottomrule[0.15em]
	\end{tabular}}}\vspace{0.5mm}
    \subfloat[\small Cross-VLM evaluation with Gemini-2.5-pro \label{tab:cross-vlm}]{\hspace{-2.5mm}
            \renewcommand{\arraystretch}{1.13}
		\scalebox{0.625}{\begin{tabular}{l c c c c c c}
				\toprule[0.15em]
				\rowcolor{color3}  Method &~~PhyT2V~~ &~~VideoDPO~~ &~Wan2.1-14B~ &~~Veo3.1~~ &~~~Sora2~~~ &~~PhyGDPO~~\\
				\midrule[0.1em]
                Overall &0.3186 & 0.3288 &0.4119 &0.5034 &0.5373 &0.5525\\
				\bottomrule[0.15em]
	\end{tabular}}}\vspace{0mm}
    \subfloat[\small Cross-T2V-model evaluation \label{tab:cross_t2v}]{
            \renewcommand{\arraystretch}{1.13}
		\scalebox{0.625}{\begin{tabular}{l c c c c}
				\toprule[0.15em]
				\rowcolor{color3}  Method &~Vcrafter2~ & ~+ Flow-DPO~ &~+ VideoDPO~ & ~+ PhyGDPO~ \\
				\midrule[0.1em]
                Overall &0.1034 &0.1128 &0.1373 &0.1426 \\
				\bottomrule[0.15em]
	\end{tabular}}}\vspace{1mm}
	\caption{\small Ablation study on VideoPhy2~\cite{videophy2}. (a) and (b) study the components of PhyAugPipe and PhyGDPO towards better performance. (c) compares VideoDPO with PhyGDPO with the same settings. (d) studies the effect of LoRA-SR scheme. (e) and (f) are cross evaluation by changing the VLM scorer to Gemini-2.5-pro~\cite{gemini_2.5} and T2V base model to Vcrafter2~\cite{videocrafter2}.}
	\label{tab:ablations}
    \vspace{1mm}
\end{table*}

\begin{figure*}[t]
	\begin{center}
		\begin{tabular}[t]{c}  \hspace{-5mm}
\includegraphics[width=1.02\textwidth]{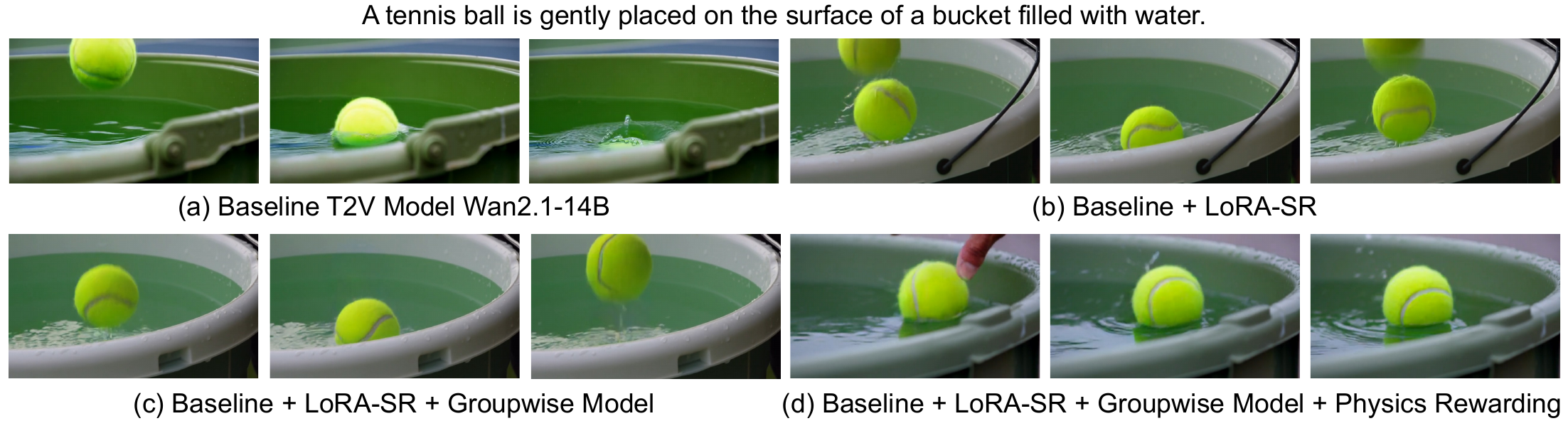}
		\end{tabular}
	\end{center}
	\vspace{1mm}
	\caption{\small Break-down visual analysis of PhyGDPO. When we progressively apply LoRA-SR, groupwise DPO modeling, and physics rewarding to Wan2.1-14B, the tennis ball in the generated video increasingly conforms to the physical law of buoyancy, floating stably on the water surface instead of sinking down or bouncing unrealistically above it.}
	\label{fig:visual_analysis}
	\vspace{1mm}
\end{figure*}

\noindent\textbf{Implementation Details.} We implement our PhyGDPO by pytorch~\cite{pytorch} based on Wan2.1-14B. Our model is finetuned for 10K steps in total at a batch size of 8 on 8 H100 GPUs for 6 days. To save GPU memory, we adopt mixed-precision training~\cite{amp} with BF16 and sublinear memory training~\cite{sublinear}. We adopt the AdamW optimizer~\cite{adamw} ($\beta_1 = 0.9$, $\beta_2 = 0.999$) with a weight decay of 0.01. The learning rate is initially set as 1e$^{-5}$ and decays to 1e$^{-6}$ using cosine annealing~\cite{cosine} algorithm. The training and inference spatial resolution of the video is 480$\times$832. We set $\tau = 3$, $N = 20000$, $\alpha_\text{min} = 0.5$, $k_\gamma = 2.0$, $b_\gamma = 0.4$, $\lambda = 0.6$, $k_\alpha = 5.0$, $b_\alpha = 0.5$. The rank and scale factor are 48.

\subsection{Comparison with State-of-the-Art Methods}

\noindent\textbf{Quantitative Results.} We compare PhyGDPO with state-of-the-art (SOTA) methods in Tab.~\ref{tab:videophy2} and~\ref{tab:phygenbench}. PhyGDPO outperforms SOTA methods by large margins. \textbf{(i)} Our method surpasses the two recent strongest closed-source models, Sora2~\cite{sora} and Veo3~\cite{veo3} across the tracks of hard actions, activities and sports, and the overall score on VideoPhy2. Especially on the hard actions, PhyGDPO achieves 4.5× higher score than the base model, Wan2.1-14B, and is 29\% and 13\% higher than Sora2 and Veo3. \textbf{(ii)} Compared to the SOTA DPO algorithm for video generation (VideoDPO) and SOTA method for physically consistent video generatiion (PhyT2V), our method surpasses VideoDPO and PhyT2V and  by 200\% and 29\% on the hard action score of VideoPhy2. Our method is also 22\%/15\% and 35\%/23\% higher than PhyT2V/VideoDPO on the mechanics and thermal tracks of PhyGenBench.

\noindent\textbf{User Study.} Since the auto-evaluation tools of VideoPhy2 and PhyGenBench are based on VLMs, their assessment of physical plausibility may not be perfect. Thus, we conduct a user study with 104 participants. Each participant is asked to pick up the video that better follows the physical laws. Tab.~\ref{tab:user_study} reports the user preference (\%) of our method over competing entries. Each participant completes 48 trials (one prompt per trial), with 24 prompts randomly sampled from VideoPhy2 and 24 from PhyGenBench. Each trial randomly selects one of the 8 baselines (non-repeated) in Tab.~\ref{tab:user_study} and compares it to our method given the same prompt. Video results are presented anonymously in random order. All results in Tab.~\ref{tab:user_study} have 95\% confidence intervals no wider than $\pm 2.4\%$. PhyGDPO is preferred by human evaluators, indicating that PhyGDPO learns genuine physics beyond VLM agreement.

\begin{table*}[t]
    \subfloat[\small Analysis of $\alpha_{\text{min}}$ \label{tab:alpha}]
    {\hspace{-3mm}
            \renewcommand{\arraystretch}{1.13}
		\scalebox{0.53}{\begin{tabular}{l c c c c c}
				\toprule[0.15em]
				\rowcolor{color3}  $\alpha_{\text{min}}$ &~~~~0.3~~~~ &~~~~0.4~~~~ &~~~~0.5~~~~ &~~~~0.6~~~~ &~~~~0.7~~~~ \\
				\midrule[0.1em]
                Overall &0.1582 & 0.1591 &0.1627 &0.1615 &0.1608 \\
				\bottomrule[0.15em]
	\end{tabular}}}\hspace{1.5mm}\vspace{0.5mm}
    \subfloat[\small Analysis of $k_{\alpha}$ \label{tab:alpha}]
    {\renewcommand{\arraystretch}{1.13}
		\scalebox{0.53}{\begin{tabular}{l c c c c c}
				\toprule[0.15em]
				\rowcolor{color3}  $k_{\alpha}$ &~~~~4.0~~~~ &~~~~4.5~~~~ &~~~~5.0~~~~ &~~~~5.5~~~~ &~~~~6.0~~~~ \\
				\midrule[0.1em]
                Overall &0.1611 &0.1579 &0.1627 &0.1593 &0.1618 \\
				\bottomrule[0.15em]
	\end{tabular}}}\hspace{1.5mm}\vspace{0.5mm}
    \subfloat[\small Analysis of $b_{\alpha}$ \label{tab:alpha}]
    {\renewcommand{\arraystretch}{1.13}
		\scalebox{0.53}{\begin{tabular}{l c c c c c}
				\toprule[0.15em]
				\rowcolor{color3}  $b_{\alpha}$ &~~~~0.3~~~~ &~~~~0.4~~~~ &~~~~0.5~~~~ &~~~~0.6~~~~ &~~~~0.7~~~~ \\
				\midrule[0.1em]
                Overall &0.1582 & 0.1579 &0.1627 &0.1598 &0.1602 \\
				\bottomrule[0.15em]
	\end{tabular}}}\vspace{0.5mm}\\
    \subfloat[\small Analysis of $\lambda$ \label{tab:alpha}]
    {\hspace{-3mm}
            \renewcommand{\arraystretch}{1.13}
		\scalebox{0.54}{\begin{tabular}{l c c c c c}
				\toprule[0.15em]
				\rowcolor{color3}  $\lambda$ &~~~~0.3~~~~ &~~~~0.4~~~~ &~~~~0.5~~~~ &~~~~0.6~~~~ &~~~~0.7~~~~ \\
				\midrule[0.1em]
                Overall &0.1592 & 0.1586 &0.1608 &0.1627 &0.1615 \\
				\bottomrule[0.15em]
	\end{tabular}}}
    \subfloat[\small Analysis of $k_{\gamma}$ \label{tab:alpha}]
    {\renewcommand{\arraystretch}{1.13}
		\scalebox{0.54}{\begin{tabular}{l c c c c c}
				\toprule[0.15em]
				\rowcolor{color3}  $k_{\gamma}$ &~~~~1.0~~~~ &~~~~1.5~~~~ &~~~~2.0~~~~ &~~~~2.5~~~~ &~~~~3.0~~~~ \\
				\midrule[0.1em]
                Overall &0.1588 & 0.1609 &0.1627 &0.1592 &0.1598 \\
				\bottomrule[0.15em]
	\end{tabular}}}
    \subfloat[\small Analysis of $b_{\gamma}$ \label{tab:alpha}]
    {\renewcommand{\arraystretch}{1.13}
		\scalebox{0.54}{\begin{tabular}{l c c c c c}
				\toprule[0.15em]
				\rowcolor{color3}  $b_{\gamma}$ &~~~~0.3~~~~ &~~~~0.4~~~~ &~~~~0.5~~~~ &~~~~0.6~~~~ &~~~~0.7~~~~ \\
				\midrule[0.1em]
                Overall &0.1586 & 0.1627 &0.1582 &0.1607 &0.1595 \\
				\bottomrule[0.15em]
	\end{tabular}}}
	\caption{\small Analysis of the hyperparameters in the physics rewarding of Eq.\eqref{eq:alpha_gamma} on the VideoPhy2~\cite{videophy2} benchmark. When tweaking a hyperparameter to analyze its effectiveness, other hyperparameters are fixed at their optimal values. }
    \vspace{1mm}
	\label{tab:analysis}
\end{table*}

\begin{figure*}[t]
	\begin{center}
		\begin{tabular}[t]{c}  \hspace{-5mm}
\includegraphics[width=1.02\textwidth]{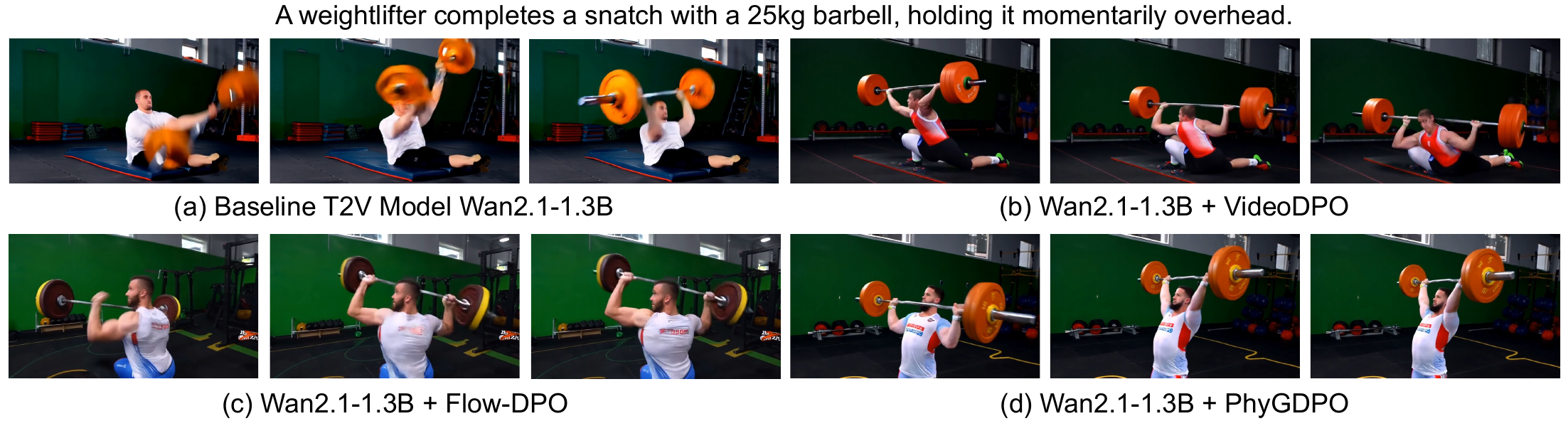}
		\end{tabular}
	\end{center}
	\vspace{1mm}
	\caption{\small Visual comparison with state-of-the-art DPO algorithms on Wan2.1-T2V-1.3B under the same training settings and data for fairness. Using our PhyGDPO can generate more physically plausible video with undistorted human body motion.}
	\label{fig:dpo_compare}
	\vspace{2mm}
\end{figure*}

\noindent\textbf{Qualitative Results.} The visual comparisons on challenging physical actions or phenomena including gymnastics, soccer, basketball, glass smashing, polo, squash, handspring,  refraction, and combustionare shown in Fig.~\ref{fig:teaser}, \ref{fig:comparison_1}, \ref{fig:comparison_2}, and \ref{fig:comparison_3}. Our method generates videos with more realistic physical dynamics, deformation-free body motion, and accurate object interactions, effectively capturing phenomena such as force transfer, material deformation, light refraction, and flame propagation. Compared to existing open- and closed-source models, PhyGDPO shows superior generalization from human activities to complex physical events. Please refer to the project page for video results.

\subsection{Results on Image-to-Video Generation}

\begin{figure*}[t]
	\begin{center}
		\begin{tabular}[t]{c}  \hspace{-7.5mm}
\includegraphics[width=1.04\textwidth]{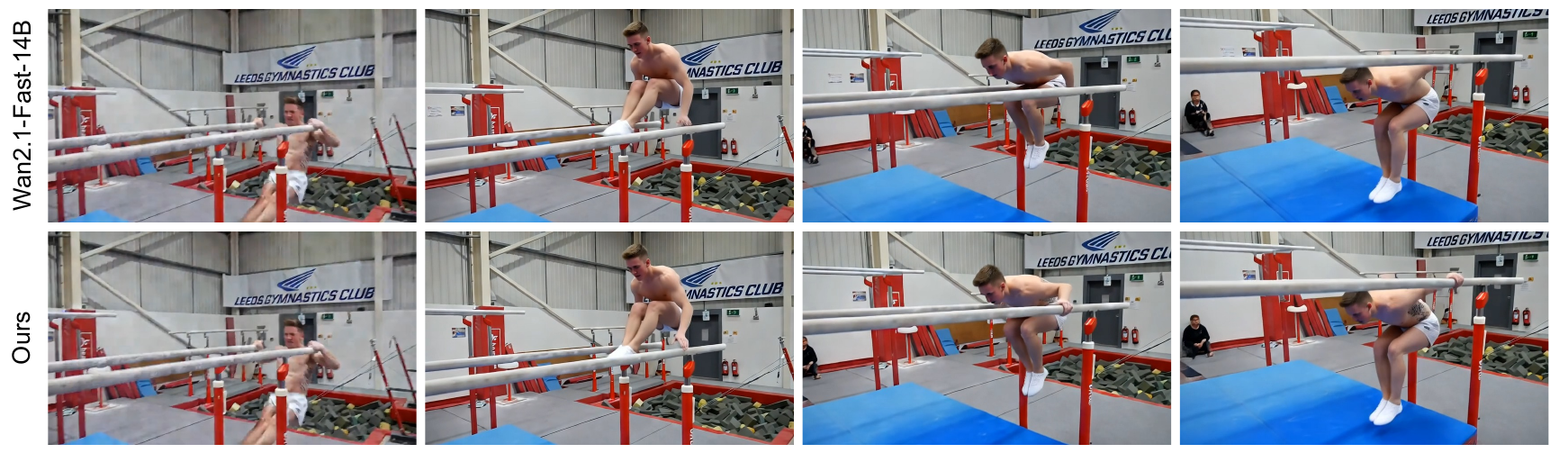}
		\end{tabular}
	\end{center}
	\caption{\small Application of PhyGDPO on image-to-video generation. When using our method to post-train the Wan2.1-Fast-14B model, the gymnast exhibits more anatomically complete body structures and smoother, more natural movements.}
	\label{fig:i2v}
\end{figure*}

We further apply our PhyGDPO to a 4-step efficient image-to-video generation model, Wan2.1-Fast-14B, to validate its generalization ability. The visual comparison on the challenging gymnastic scene is shown in Fig.~\ref{fig:i2v}. After post-training with PhyGDPO, the gymnast exhibits improved anatomical integrity and temporal consistency in both body appearance and motion. For instance, the tattoo on the left side of the gymnast's waist is consistently preserved, while both hands remain free of distortions and perform physically plausible and well-coordinated gripping and releasing motions on the bar. These results demonstrate the strong generalizability of PhyGDPO to image-to-video generation, effectively enhancing both physical plausibility and temporal consistency.

\subsection{Ablation Study}

\vspace{1mm}
\noindent\textbf{Ablation of PhyAugPipe.} We conduct a break-down ablation to study the effect of each component of PhyAugPipe in Tab.~\ref{tab:phyaugpipe}. The baseline is directly training PhyGDPO on the randomly selected text-video data. When we keep the number of selected training data the same and apply the data filtering with CoT, action clustering with semantics matching, and data sampling with physics-guided rewarding, all track scores are improved. Especially the score on the hard actions gains by over 50\%. These results suggest the effectiveness of our data construction techniques.

\vspace{1mm}
\noindent\textbf{Ablation of PhyGDPO.} Tab.~\ref{tab:phygdpo} shows a break-down ablation for PhyGDPO. When progressively using LoRA-SR, PL groupwise probabilistic model, and physics-guided rewarding, the performance steadily gains by large margins. The score on the hard action categories yields 4.5$\times$ improvement, showing the effectiveness of our designed methods in improving T2V physics plausibility. In addition, we conduct a visual analysis in Fig.~\ref{fig:visual_analysis} (a) for PhyGDPO. When gradually using our proposed techniques, the tennis in the generated video no longer presents ghosting artifacts, and its floating motion on the water surface better conforms to the physical laws of fluid buoyancy.

\vspace{1mm}
\noindent\textbf{PhyGDPO \emph{vs.} SOTA DPO.} For fair comparison with two SOTA DPO methods: Flow-DPO~\cite{flowdpo} and VideoDPO, we use them to finetune Wan2.1-1.3B with the same data and settings. As reported in Tab.~\ref{tab:dpo_compare}, PhyGDPO surpasses Flow-DPO and VideoDPO across all tracks by large margins, especially on the hard action track, where it yields 50\% improvement. We conduct a visual comparison in Fig.~\ref{fig:visual_analysis} (b) on the challenging weightlifting action. Flow-DPO and VideoDPO generate distorted or unstable body poses. In contrast, PhyGDPO generates coherent body motion with stable shapes and balanced force dynamics, highlighting its advantage in modeling complex physical actions.

\vspace{1mm}
\noindent\textbf{Analysis of LoRA-SR.} We conduct an ablation study of LoRA-SR in Tab.~\ref{tab:lora_sr}. In our PhyGDPO training, using LoRA-SR reduces GPU memory consumption by 44\% and achieves over 60× compression in storage size, while still improving the score on hard actions by 14\% compared to the version without LoRA-SR. We also conduct a visual analysis in Fig.~\ref{fig:visual_analysis} (c), using our LoRA-SR in DPO training can generate more plausible physical interactions between human hands and the snow shovel. Besides, we also compare with SFT using the same LoRA in Tab.~\ref{tab:lora_sr}. Our PhyGDPO with LoRA-SR uses almost the same GPU memory but surpasses LoRA-SFT by large margins, especially on the hard-action categories, where the score achieves 2.7$\times$ improvement.

\vspace{1mm}
\noindent\textbf{Cross Evaluation.} \textbf{(i)} To provide stronger evidence that our method improves the physical plausibility, we conduct a cross-VLM evaluation. We replace the VLM scorer of VideoPhy2 by a stronger physics evaluator, Gemini-2.5-pro~\cite{gemini_2.5}, to score the generated videos with the same protocol in Tab.~\ref{tab:cross-vlm}. Our method still outperforms SOTA T2V models, showing that our method learns better genuine physics. \textbf{(ii)} We change the T2V model to Vcrafter2~\cite{videocrafter2} and apply different DPO methods on it with the same setting for fairness in Tab.~\ref{tab:cross_t2v}. PhyGDPO consistently surpasses SOTA DPO methods, suggesting the superiority and generalization ability of our method. 

\begin{figure*}[t]
	\begin{center}
		\begin{tabular}[t]{c}  \hspace{-2.8mm}
\includegraphics[width=1.01\textwidth]{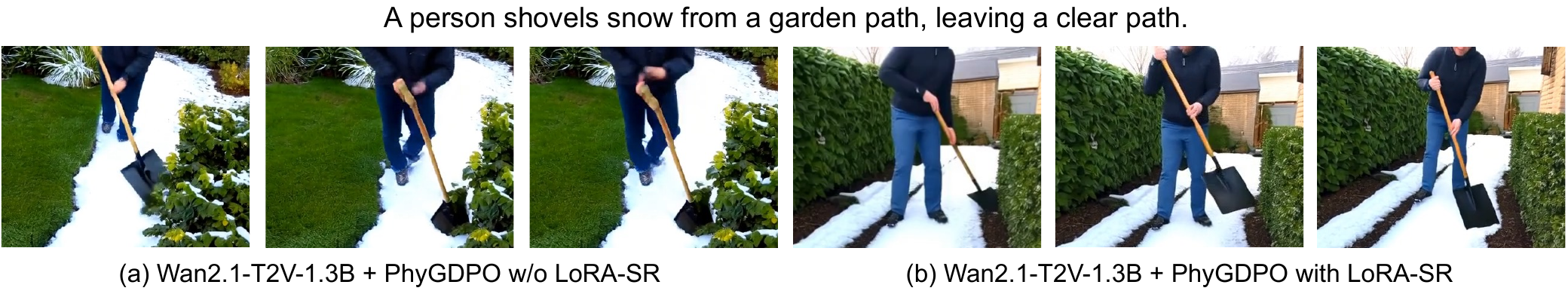}
		\end{tabular}
	\end{center}
	\caption{\small Visual analysis of our LoRA-SR scheme.  Applying our LoRA-SR can generate more accurate and plausible hand–shovel interaction in the snow shoveling scenario. In contrast, removing LoRA-SR generates distorted hand actions.}
    \vspace{2mm}
	\label{fig:visual_analysis}
\end{figure*}

\noindent\textbf{Parameter Analysis.} We analyze the effects of the hyperparameters in Eq.~\eqref{eq:alpha_gamma}. When tweaking a hyperparameter, other hyperparameters are fixed at their optimal values. The results are shown in Tab.~\ref{tab:analysis}. The best performance is achieved when $\alpha_{\text{min}} = 0.5$, $k_\alpha = 5.0$, $b_\alpha = 0.5$, $\lambda = 0.6$, $k_\gamma = 2.0$, and $b_\gamma = 0.4$. As shown in Tab.~\ref{tab:phyaugpipe}, when we do not use the physics-guided rewarding, the model achieves 0.1575 on the overall score. And across all hyperparameter configurations of Tab.~\ref{tab:analysis}, our method consistently achieves performance gains.

%% file: sec/05_related_work.tex
\section{Related Work}
\label{section:related_work}

\subsection{Text-to-Video Generation Models}
T2V generation~\cite{sora,omnipaint,ho2022imagen,magicvideo,svd,he2022latent,svd,flow_matching_1,pyramid_flow,ho2022video,i2vgen_xl,lavie} has witnessed significant progress. Many existing works follow DiT~\cite{dit} to adopt a Transformer~\cite{transformer} to predict the noise in diffusion~\cite{snapvideo,li2023videogen,vdt,ma2024latte,Lumina_t2x,moviegen,diffusiongs,gupta2024photorealistic,videopoet} or estimate the velocity field in flow matching~\cite{wan,pyramid_flow,flow_matching_1,gao2025seedance,omnivcus,editverse,zhang2025waver,wang2025universe,flashvideo,flow_matching_4,goku}. Although high visual quality is achieved, accurately modeling the underlying physics-related effects remains challenging. Two recent works, DiffPhy~\cite{diffphy} and PhyT2V~\cite{phyt2v}, adopt an LLM to extend the text prompt with explicit physics laws and phenomena, and then iteratively generates the video or finetunes a T2V model with extended prompts. Yet, these prompt extension-based methods are easily misled by the mistakes of LLM and struggle to learn implicit physics. Different from these previous works, our goal is to explore the physics reasoning potential of the video generation model itself.

\subsection{Direct Preference Optimization}
Unlike reinforcement learning from human feedback (RLHF), which requires training a separate reward model and performing reinforcement learning, Direct Preference Optimization (DPO)~\cite{dpo} directly learns from preference pairs in a closed-form objective, making the training pipeline simpler and more stable.
 In recent years, DPO has been adopted in image~\cite{diffusiondpo,image_dpo_1,image_dpo_2,image_dpo_3} and video~\cite{videodpo,physhpo,flowdpo,densedpo,ipo} generation. For example, DiffusionDPO~\cite{diffusiondpo} finetunes a text-to-image diffusion model with human-annotated preference image pairs. VideoDPO~\cite{videodpo} adapts DiffusionDPO into T2V generation. However, these DPO frameworks mainly improve the visual quality and aesthetics. 
In addition, these DPO methods suffer from a severe low-efficiency problem because they need to copy a full model as the reference, which occupies redundant GPU memory. Our work aims to fill these research gaps.

%% file: sec/06_conclusion.tex
\section{Conclusion}
In this paper, we focus on studying a challenging problem, physically consistent T2V generation, without using LLM for prompt extension in inference because our goal is to explore the implicit physics reasoning ability of the video generation model. To this end, we first propose a data construction method, PhyAugPipe, to collect a training dataset PhyVidGen-135K containing over 135K text-video data pairs with rich physics interaction and phenomena. Subsequently, we formulate a principled groupwise DPO framework, PhyGDPO, with two technical designs PGR and LoRA-SR. PhyGDPO efficiently post-trains Wan2.1-T2V-14B model on PhyVidGen-135K to boost the physics plausibility in video generation. Comprehenseiv experiments show that our method quantitatively and qualitatively outperforms SOTA algorithms and achieves higher human preference in the user study. The application of PhyGDPO on image-to-video generation also suggests the generalization ability of our algorithm.

%% file: sec/XX_supp.tex
\clearpage
\newpage
\beginappendix

\section*{Mathematical proof for Ineq.~(\textcolor{metablue}{8})}

In the main paper, we use an inequality (\textcolor{metablue}{8}) as
\vspace{-1mm}
\begin{equation}
\small
1+\sum_{j=1}^m e^{x_j} \leq
\prod_{j=1}^m \big(1+e^{\alpha_j x_j}\big)^{\gamma_j}, ~
 ~~
0 < \alpha_j \leq 1,\ \gamma_j \geq 1/\alpha_j.
\label{eq:prod-ineq_sup}
\vspace{-1mm}
\end{equation}
Here we prove it. For each $j$, since $0 < \alpha_j \le 1$, the function $f_j(t)=t^{\alpha_j}$ is concave on $[0,+\infty)$. 
For any $u,v\ge 0$, let $s = u+v$. If $s=0$ the inequality is trivial. 
Assume $s>0$ and set $\lambda = \frac{u}{s}\in[0,1]$, so that $u=\lambda s$ and $v=(1-\lambda)s$. 
By the concavity of $f_j$, we can derive the following inequality:
\begin{equation}
\begin{aligned}
    f_j(u) &= f_j(\lambda s + (1-\lambda)\cdot 0) \\
&\ge \lambda f_j(s) + (1-\lambda) f_j(0)
= \lambda f_j(s),
\label{eq:u_1}
\end{aligned}
\end{equation}
and similarly, we derive the inequality for $f_j(v)$ as
\begin{equation}
f_j(v) = f_j\big((1-\lambda) s + \lambda\cdot 0\big)
\;\ge\;
(1-\lambda) f_j(s).
\label{eq:v_1}
\end{equation}
Adding the two inequalities Ineq.~\eqref{eq:u_1} and Ineq.~\eqref{eq:v_1} yields
\begin{equation}
\begin{aligned}
u^{\alpha_j} + v^{\alpha_j}
&= f_j(u) + f_j(v) \\
&\ge \lambda f_j(s) + (1-\lambda) f_j(s) \\
& = f_j(s) = (u+v)^{\alpha_j},
\end{aligned}
\end{equation}
this is,
\begin{equation}
(u+v)^{\alpha_j} \le u^{\alpha_j} + v^{\alpha_j}.
\label{eq:concave-alpha}
\end{equation}
Applying Ineq.~\eqref{eq:concave-alpha} to $u=1$ and $v=e^{x_j}$ yields
\begin{equation}
\big(1+e^{x_j}\big)^{\alpha_j} \le 1 + e^{\alpha_j x_j},
\end{equation}
which is equivalent to
\begin{equation}
1 + e^{x_j} \le \big(1+e^{\alpha_j x_j}\big)^{1/\alpha_j}.
\label{eq:softplus-bound}
\end{equation}
Since $\gamma_j \ge 1/\alpha_j$ and the base $\big(1+e^{\alpha_j x_j}\big)\ge 1$, we have
\begin{equation}
\big(1+e^{\alpha_j x_j}\big)^{1/\alpha_j}
\le \big(1+e^{\alpha_j x_j}\big)^{\gamma_j},
\end{equation}
and thus from Ineq.~\eqref{eq:softplus-bound}, we can easily derive:
\begin{equation}
1 + e^{x_j} \le \big(1+e^{\alpha_j x_j}\big)^{\gamma_j}
\Longrightarrow
e^{x_j} \le \big(1+e^{\alpha_j x_j}\big)^{\gamma_j} - 1.
\label{eq:single-bound}
\end{equation}
We define $a_j \triangleq \big(1+e^{\alpha_j x_j}\big)^{\gamma_j} - 1 \ge 0$. Then Ineq.~\eqref{eq:single-bound} gives $e^{x_j} \le a_j$ for all $j$, and hence
\begin{equation}
\sum_{j=1}^m e^{x_j} \;\le\; \sum_{j=1}^m a_j.
\label{eq:sum-a}
\end{equation}
On the other hand, for any nonnegative $\{a_j\}_{j=1}^m$ we have
\begin{equation}
\small
\prod_{j=1}^m (1+a_j)
= 1 + \sum_{j=1}^m a_j + \text{(higher-order terms in $\{a_j\}$)},
\end{equation}
where all higher-order terms are nonnegative, so
\begin{equation}
1 + \sum_{j=1}^m a_j \;\le\; \prod_{j=1}^m (1+a_j).
\label{eq:sum-prod}
\end{equation}
Combining Ineq.~\eqref{eq:sum-a} and Ineq.~\eqref{eq:sum-prod} and substituting the definition of $a_j$ yields
\begin{equation}
\begin{aligned}
1 + \sum_{j=1}^m e^{x_j}
&\le
1 + \sum_{j=1}^m a_j
\le
\prod_{j=1}^m (1+a_j) \\
&=
\prod_{j=1}^m \big(1 + \big(1+e^{\alpha_j x_j}\big)^{\gamma_j} - 1\big) \\
&=
\prod_{j=1}^m \big(1+e^{\alpha_j x_j}\big)^{\gamma_j},
\end{aligned}
\end{equation}
which is exactly the claimed Ineq~\eqref{eq:prod-ineq} used in the main paper.